\documentclass{article}
\usepackage{ijcai22}
\usepackage{xcolor}

\usepackage{amssymb} 
\usepackage{amsmath} 
\usepackage{array} 
\newcolumntype{P}[1]{>{\centering\arraybackslash}p{#1}} 
\usepackage{booktabs} 
\usepackage{multirow} 
\usepackage{subfigure} 
\usepackage{paralist}
\usepackage{times}
\usepackage{soul}
\usepackage{url}
\usepackage[hidelinks]{hyperref}
\usepackage[utf8]{inputenc}
\usepackage[small]{caption}
\usepackage{graphicx}
\usepackage{amsmath}
\usepackage{amsthm}
\usepackage{booktabs}
\usepackage{algorithm}
\usepackage{algorithmic}
\title{RAW-GNN: RAndom Walk Aggregation based Graph Neural Network}

\author{
	Anonymous Author(s)
	\affiliations
	\emails
}

\author{
	Di Jin$^1$\and
	Rui Wang$^1$\and
	Meng Ge$^1$\footnote{Corresponding authors}\and
	Dongxiao He$^{1*}$\and
	Xiang Li$^2$\and
	Wei Lin$^2$\And
	Weixiong Zhang$^3$
	\affiliations
	$^1$College of Intelligence and Computing, Tianjin University, Tianjin, China\\
	$^2$Meituan, Beijing, China\\
	$^3$Department of Health Technology and Informatics, The Hong Kong Polytechnic University, Kowloon, Hong Kong
	\emails
	\{jindi, wr1895, gemeng, hedongxiao\}@tju.edu.cn,
	weixiong.zhang@polyu.edu.hk
}

\begin{document}
	
	\maketitle
	
	\begin{abstract}
	    Graph-Convolution-based methods have been successfully applied to representation learning on homophily graphs where nodes with the same label or similar attributes tend to connect with one another. Due to the homophily assumption of Graph Convolutional Networks (GCNs) that these methods use, they are not suitable for heterophily graphs where nodes with different labels or dissimilar attributes tend to be adjacent. Several methods have attempted to address this heterophily problem, but they do not change the fundamental aggregation mechanism of GCNs because they rely on summation operators to aggregate information from neighboring nodes, which is implicitly subject to the homophily assumption. Here, we introduce a novel aggregation mechanism and develop a RAndom Walk Aggregation-based Graph Neural Network (called RAW-GNN) method. The proposed approach integrates the random walk strategy with graph neural networks. The new method utilizes breadth-first random walk search to capture homophily information and depth-first search to collect heterophily information. It replaces the conventional neighborhoods with path-based neighborhoods and introduces a new path-based aggregator based on Recurrent Neural Networks. These designs make RAW-GNN suitable for both homophily and heterophily graphs. Extensive experimental results showed that the new method achieved state-of-the-art performance on a variety of homophily and heterophily graphs.
	\end{abstract}
	
	\section{Introduction}
	Graphs are ubiquitous in the real world, such as social networks, brain networks, transportation networks and citation networks. Network analysis \cite{networkanalysis1,JINDI01} has been extensively studied and broadly applied in many fields, ranging from computer science to social sciences, biology, physics, and many more. Recently, message-passing neural networks (MPNNs) \cite{GCN} have been successfully adopted for various problems on graphs, e.g., node classification, graph classification, link prediction, and anomaly detection \cite{HOGGCN,JINDI02}.
	
	An MPNN runs an iterative process and in each iteration, every node sends its features as a message to its neighbors and then aggregates messages from other nodes to update its representation. GCN \cite{GCN}, as a typical MPNN, works under the homophily assumption – i.e., most connected nodes are from the same class and have similar attributes. GCN and its variants, like GraphSAGE \cite{GraphSage}, possess great power for learning on graphs and have shown excellent performance on many downstream tasks on networks with homophily.
	
	
	However, many real-world networks do not satisfy the homophily assumption. But rather, there exist many heterophily or low homophily networks where most adjacent nodes may belong to different classes and have dissimilar attributes. For example, in protein structures, an amino acid type is more likely to connect to different amino acid types rather than the same amino acid type; In email networks, spam users often contact normal users; In e-commerce networks, fraudsters are more likely to connect to accomplices than to other fraudsters. Conventional GCNs are not designed for heterophily networks and they use message propagation mechanisms based on the homophily assumption, as a result, information from different classes will get mixed during propagation on heterophily networks using such message propagation methods.
	
	
	To deal with the homogeneity restriction in GCNs, several methods have already been proposed. Based on the aggregation mechanisms that they use, these methods can be divided into two categories:
	(1) Methods adjusting attention weights between neighbors of different labels in aggregation, which include GGCN \cite{GGCN}, CPGNN \cite{CPGNN}, HOG-GCN \cite{HOGGCN}, GNN-LF/HF\cite{WangXiaoWWW21} and BM-GCN \cite{BMGCN}. In essence, these methods assign a weight to each connected node pair with the guidance of heterophily information. In specific, label-guided methods \cite{CPGNN,HOGGCN,BMGCN} integrate label information into their framework and aggregate nodes with the same label in the neighborhood and reduce the degree of aggregation of neighbors with different labels; Signed-message-guided methods \cite{GGCN,WangXiaoWWW21} extend attention weight from $[0, 1]$ to $[-1, 1]$. As a result, messages between nodes of the same type of label are assigned positive weights, which boost message passing in the same class, and messages between nodes of different labels are given negative values, which prevent dissimilar neighbors from passing harmful and irrelevant information to one another.
	(2) Methods directly aggregating messages among higher-order neighbor nodes, including Geom-GCN \cite{Geom-GCN}, H2GCN \cite{H2GCN}, GPR-GNN \cite{GPR-GNN} and LINKX \cite{LINKX}. These methods assume that directly adjacent neighborhoods may be heterophily-dominant for heterophily network, but the higher-order neighborhoods are homophily-dominant which can provide more useful information for the target node. By explicitly aggregating information from higher-order neighborhoods, these methods alleviate the heterophily problem to certain extent.

	However, these methods do not change the fundamental aggregation mechanism of GCNs because of the summation operator in the aggregation process, which is implicitly subject to the homophily assumption. To be specific, label-guided methods try to learn the weights between nodes of different labels to $0$. In fact, this is equivalent to omitting the information of heterophily nodes in the aggregation, which is useful for network representation learning and downstream tasks. In signed-message-guided methods, weight $-1$ is introduced based on an analysis of graphs with two kinds of labels, so that this design is only effective on networks with a small class number close to $2$ and do not work well for networks with large number of classes. In short, these designs alleviate the homogeneity restriction problem to some extent, but they are still constraint by the summation aggregator. In addition, the existing methods that directly aggregate higher-order neighbor use the classical summation aggregator to aggregate higher-order neighbors based on their distances to the target node and concatenate aggregated results with different distances. These methods also suffer from the constraints of the summation aggregator.
	
	
	%
	To address these problems, we introduce a new aggregation mechanism and propose a RAndom Walk aggregation-based Graph Neural Network, short-handed as RAW-GNN. We integrate random walk sampling into graph neural networks and extend the conventional neighborhoods to $k$-hop path-based neighborhoods. A $k$-hop path formed by random walks preserves the original attributes on this $k$ nodes and the original structural connections of these nodes in the random walk sequence. In this way, the path-based neighborhoods can represent the neighborhood distribution of the target node better than the conventional neighborhoods. Furthermore, we utilize breadth-first search random walk (BFS) to capture homophily information and depth-first search (DFS) to collect heterophily information. To go beyond the exiting aggregation mechanism of GCNs and to take full advantage of the path-based neighborhoods, we adopt a sequential Recurrent Neural Networks (RNN) based aggregator, which can take into consideration the order information of neighbor nodes preserved by random walks. The RNNs have the advantage that they can handle the diverse attributes of adjacent nodes, which accommodates the need for heterophily networks. Finally, we use the attention mechanism to learn the importance of different paths from DFS channel (and BFS channel respectively), which can better extract heterophily (and homophily) neighborhood distribution with minimal mixing of information and can enable the model to automatically make a trade-off between homophily and heterophily according to different network characteristics.

	\begin{figure}[t]
		\centering
		\includegraphics[width=\linewidth]{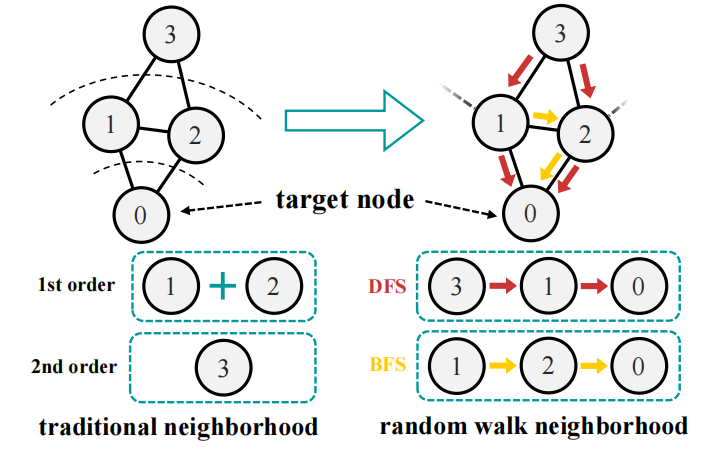}
		\caption{Existing methods omit some of the direct message channels between nodes of different distances to the target node, like edge $e_{1,3}$ and $e_{2,3}$.
			RAW-GNN employs the path-based neighborhoods detected by random walks to tackel this problem.}
		\label{figure_example}
	\end{figure}

	\section{Preliminaries}
	We now present notations and related definitions, including path-based neighborhood and generalized homophily ratio, which are important for our work.
	\subsection{Basic Notations}
	Let $G = (V, E, X)$ denote an undirected, unweighted and attributed network, where $V = \left \{v_1,v_2, . . . ,v_n \right \}$ is a set of $n$ nodes and $X \in \mathbb{R}^{n \times f}$ is a set of node attributes. Every node $i$ is associated with $f$ attributes $x_i$, which form the $i$-th row of $X$. $E = \left \{e_{ij} \right \} \subseteq V  \times V$ is a set of edges represented by an adjacency matrix $A=[a_{ij}] \in \mathbb{R}^{n \times n}$, where $a_{ij}=1$ if nodes $v_i$ and $v_j$ are connected, or $a_{ij} = 0$ otherwise.
	
	In this paper, we focus on semi-supervised node classification task. In a semi-supervised node classification task, every node belongs to a class $c \in C$ and $|C|$ is the number of classes. Here, the labels of nodes are given in set $V_L$. Every node $v_i \in V_L$ is assigned with a label $y_i \in L = \left \{1, 2,...|C|\right \}$. The objective of node classification task is to predict the labels for all the unlabeled nodes in node set $V\backslash {V_L}$.
	
	\subsection{Message-Passing Neural Network}
	In a message-passing neural network, neighborhood is defined as neighbor nodes that are one or more hops away. Messages from nearby nodes are aggregated to the target node and updated iteratively. The $l$-th layer of a MPNN can be defined as follows:
	
	\begin{equation} \label{equation_mpnn}\begin{split}
			n_i^{(l)} =& \operatorname{aggregate}^{(l)}\left(\left\{h_i^{(l-1)}:i \in N_i\right\}\right)\\
			h_i^{(l)} =& \operatorname{combine}^{(l)}\left(h_i^{(l-1)},n_i^{(l)}\right)
		\end{split} 
	\end{equation}
	where $h_i^{(l)}$ is the feature vector of node $i$ in the $l$-th layer. The beginning vector $h_i^{(0)}$ is $x_i$, and $N_i$ is a set of neighbor nodes of node $i$. Different choices of $N_i$, $\operatorname{aggregate}^{(l)}$ and $\operatorname{combine}^{(l)}$ result in different models. Usually, the neighbor nodes of node $i$ are either the adjacent nodes of $i$ or the $k$-hop neighbor nodes of $i$. In the following subsection, we will define another kind of neighborhood, which is based on paths.
	
	\subsection{Path-based Neighborhoods} The path-based neighborhood has already been applied to heterogeneous graphs \cite{MAGNN} and knowledge graphs \cite{CogKR}. Here we modify their definition to make it suitable for heterophily graphs. Formally, a path $P$ is defined in the form of an ordered list $P = \left \{v_{p_1},v_{p_2}, . . . ,v_{p_K} \right \}$, where $v_{p_k} \in V, k=1,2,...,K$ and $e_{p_kp_{k+1}} \in E, k=1,2,...,K-1$, and $K$ is the length of the path. A path-based neighbor of node $i$ is denoted as $P_i$, where the ending node in the list $v_{p_K}$ is $v_i$. All path-based neighbor $P_i$ of node $i$ collected by strategy $s \in S$ forms the path-based neighborhood $N_i^{s}$, where $S$ is the set of strategies. In other words, all $P_i \in N_i^{s}$.
	
	\begin{figure*}[ht] 
		\centering
		\includegraphics[width=\linewidth]{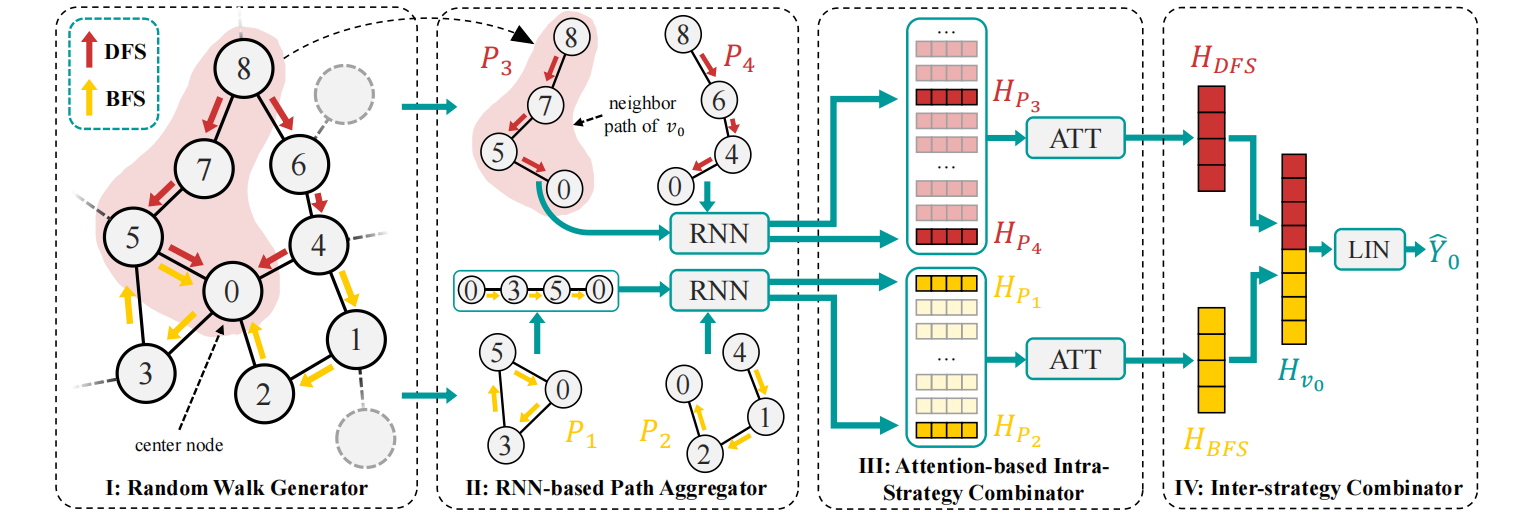}
		\caption{The framework of the proposed RAW-GNN. We extend the  neighborhood of traditional GCNs to the ordered path-based neighborhood sampled by the random walk generator. Then the sequential RNN-based aggregator is applied. This combination can alleviate the problem brought by the difference of node attributes of adjacent nodes under heterophily. Next, the attention-based intra-strategy combinator receives path embeddings sampled by the same strategy and combines them into a strategy-specific embedding. Last, the inter-strategy combinator concatenates embeddings from different strategies to get the final embedding with the minimal information mixing.
		}
		\label{theModel}
	\end{figure*}
	
	\subsection{Generalized Edge Homophily Ratio}
	The edge homophily ratio \cite{H2GCN} measures the overall homophily in a graph. Specifically, it measures the fraction of edges that connect nodes that have the same label and it is defined as $H_E^{label}(G) = {|\{ (u,v):e_{u,v} \in E \wedge {y_u} = {y_v}\} |} / {{|E|}}$. This metric is based on node labels. Here we generalize edge homophily ratio to node features. First, we define the similarity function between two nodes as $\operatorname{sim}: V \times V \to [0,1]$. This function should have such property that, when node $i$ and node $j$ are similar,  $\operatorname{sim}(i,j)$ is close to $1$ and when node $i$ and node $j$ are dissimilar, $\operatorname{sim}(i,j)$ is close to $0$. Then the generalized edge homophily ratio $H_E:G \to [0,1]$ is defined as:
	\begin{equation} \label{GeneralizedHomophilyRatio}
		\begin{split}
			{H_E}(G) = \frac{{\sum\nolimits_{(i,j) \in E} {\operatorname{sim}(i,j)} }}{{\left| E \right|}}
		\end{split} 
	\end{equation}
	when the similarity function $\operatorname{sim}(i,j)$ is Eq. (\ref{equation_sim}), Eq. (\ref{GeneralizedHomophilyRatio}) becomes the label-based edge homophily ratio mentioned above. 
	\begin{equation} \label{equation_sim}
		\begin{split}
			\operatorname{sim}(i,j) = \left\{ {\begin{array}{*{20}{c}}
					{1,{y_i} = {y_j}} \\ 
					{0,{y_i} \ne {y_j}} 
			\end{array}} \right.
		\end{split} 
	\end{equation}
	We can also use node features to define node similarity, e.g., cosine similarity in Eq. (\ref{equation_cos}), then we can generalize the concept of homophily ratio to features.
	
	\begin{equation}  \label{equation_cos}
	\operatorname{sim}(i,j) = \operatorname{cos}(x_i,x_j)
	\end{equation}
	
	
	
	\section{The Architecture}
	In this section, we describe the proposed RAW-GNN for homophily and heterophily graph embedding. We start with a brief overview and then introduce the details of components.
	
	\subsection{Overview}

   To let the aggregation mechanism truly go beyond the homophily assumption, here we propose a novel approach that consists of four components: random walk generator, RNN-based path aggregator, attention-based intra-strategy combinator and inter-strategy combinator. The whole framework of our approach RAW-GNN is shown in Fig. \ref{theModel}.
    Different from the frameworks of other GCNs, we encode the homophily and heterophily information into two different channels. In specific, we use two random walk generators to sample $k$-hop paths, i.e., using Breadth-First Search random walk (BFS) generator to sample homophily neighborhood distribution and Depth-First Search (DFS)  generator to sample heterophily neighborhood distribution. Each sampling strategy (BFS or DFS) will sample multiple paths so as to capture the neighborhood distribution more accurately. After getting BFS neighborhood paths and DFS neighborhood paths, two RNN-based path aggregators are used to aggregate the ordered attributes of nodes in a path to form a path embedding. Then, we use the attention mechanism to learn the importance of different paths from DFS strategy and BFS strategy respectively, and form two strategy-specific embeddings. At last, we concatenate the two embeddings from BFS channel and DFS channel to form the whole embedding which can represent homophily and heterophily information together in the same framework.


	\subsection{Random Walk Generator}
	The key role of neighbors is to provide useful information for the target node, so it doesn't have to be like neighborhoods in classical GCNs. Here, we use paths to define neighborhood. However, given a receptive field, there are more paths than nodes so that it is not computationally feasible to use all the paths especially when the receptive field is large. To collect more information with fewer paths in the search space, we adopt a $2$-nd order random walk with two parameters $p$ and $q$ from  Node2Vec \cite{Node2Vec} to simulate Breadth-first Search and Depth-first Search.
	
	Consider a random walk that has traversed edge $e_{t,s}$ , is now on node $s$ and is going to visit the next node $r$. We set the unnormalized transition probability as follows:
	\begin{equation} \label{equation_random_walk}
		{P_{pq}}\left( {{v_i=r}|{v_{i - 1}} = s,{\text{ }}{{{v}}_{i - 2}} = t} \right) = \left\{ {\begin{array}{*{20}{c}}
				{1/p{\text{ }}\operatorname{if}{\text{ }}{d_{tr}} = 0} \\ 
				{1{\text{ }}\operatorname{if}{\text{ }}{d_{tr}} = 1} \\ 
				{1/q{\text{ }}\operatorname{if}{\text{ }}{d_{tr}} = 2} \\ 
				{0{\text{   }}otherwise} 
		\end{array}} \right.
	\end{equation}
	where $d_{tr}$ is the length of the shortest path between nodes $s$ and $r$. When we set $p<1$ and $q>1$, the random walk tends to behave like BFS; and when $p>1$ and $q<1$, the random walk tends to behave like DFS. 
	
	BFS collects information from the immediate neighbors of the target node and can extract homophily information (Fig. \ref{theModel}). DFS tends to sequentially reach nodes at increasing distances from the source node, which can extract heterophily information. Since real networks exhibit both homophily and heterophily, so both search strategies are applied.
	
	\subsection{RNN-based Path Aggregator}
	Given a path $P_i$, a path aggregator should learn the structural and semantic information of all nodes on $P$, not just the starting node $v_i$ and the ending target node, but also all the context nodes in between. Note that the existing methods that treat all neighboring nodes as an unordered set. Nevertheless, a path naturally comes with an order, which also preserves the ordered connections among the nodes on the path. The objective of the path aggregator is to encode all the nodes on the path and consider the order of the nodes on the path to preserve more relational information among the connected nodes. The path aggregator $\operatorname{f_{path}}^{(l)}: \mathbb{R}^{K \times d_l} \to \mathbb{R}^{d_{l}}$ for the $l$-th layer is defined as:
	\begin{equation} \label{equation_PathAgg}
		h_P^{(l)} = \operatorname{f_{path}}^{(l)}(P) = \operatorname{f_{\theta}}(\left \{h_{n_1}^{(l)},h_{n_2}^{(l)}, . . . ,h_{n_K}^{(l)} \right \})
	\end{equation}
	where $h_P^{(l)} \in \mathbb{R}^{d_{l}}$ is the $l$-th layer embedding vector of path $P = \left \{v_{n_1},v_{n_2}, . . . ,v_{n_K} \right \}$, $h_{p_k}^{(l)} \in \mathbb{R}^{d_l}$ is the feature of node $v_{n_k}$ in layer $l$, $k=1,2,...,K$ and $\theta$ represents all the learnable parameters of the aggregator. To encode a path, we can use any encoder that takes the order of elements into consideration. Here we choose a simple sequence encoders GRU (Gate Recurrent Unit) \cite{GRU}, a variant of RNN with a gating mechanism as the message function.
	
	For simplicity, we set the layer to $l$ and omit the superscript indicating the layer of an embedding in the rest of the paper unless otherwise specified.
	
	\subsection{Attention-based Intra-Strategy Combinator}
	After computing the path embeddings $h_P^s \in N_i^{S}$ for every node $v_i \in V$ for strategy $s$, we need to combine them into strategy-specific node embeddings. We assume that given a certain sampling strategy, the generated paths obey a corresponding distribution . Since all paths are sampled from the same distribution, it is reasonable to sum these paths to  approximate the corresponding distribution. It is worth noting that this summation is different from combining node embeddings in GCNs because the pattern exhibited in a path cannot be well captured in a single node and beginning with summing node embeddings like typical GCNs weakens such path pattern. Furthermore, different paths may contribute differently to the target node embedding, so that we adopt an attention mechanism to learn the different weight of the path embeddings in $N_i^{S}$, which is defined as:
	\begin{equation} \label{equation_att}
		\begin{split}
			{{\text{e}}_P} =& \operatorname{LeakyReLU}(a_P^T \cdot {h_P^s}) \hfill \\
			{\alpha _P} =& \frac{{\exp ({{\text{e}}_P})}}{{\sum\nolimits_{Q \in N_i^{S}} {\exp ({{\text{e}}_Q})} }} \hfill \\
			{h_i^s} =& \sigma (\sum\nolimits_{P \in N_i^{S}} {{\alpha _P}}  \cdot {h_P^s}) \hfill \\ 
		\end{split} 
	\end{equation}
	where $a_P \in \mathbb{R}^{d_l}$ is the learnable attention parameter, $a_P^T$ denotes the transpose of $a_P$, ${\text{e}}_P$ is the unnormalized importance weight of path $P$, and $\alpha _P$ is the path weight normalized across all the paths in $N_i^{S}$ using softmax. It is worth noting that we do not consider the embedding of the target node $h_i$ separately again during the combination procedure like Eq. (\ref{equation_mpnn}), since every path in $N_i^{S}$ already contains $h_i$ as its ending node embedding. To further stabilize the learning process, we adopt the standard multi-head attention.
	Specifically, $H$ attention heads in Eq. (\ref{equation_att}) are used and then their embeddings are concatenated, which is formalized as:
	\begin{equation} \label{equation_multi_head}
		{h_i^s} = \mathop {||}\limits_{h = 1}^H \sigma (\sum\nolimits_{P \in N_i^S} {\alpha _P^h}  \cdot {h_P^s})
	\end{equation}
	\subsection{Inter-Strategy Combinator}
	After aggregating the path information within every strategy, we need to combine them using an inter-strategy combination layer. Since different strategies gather paths of different distribution, in order to combine embeddings from different strategies without mixing them, we use concatenation to combine embeddings from different strategies, rather than summing them as done in the GCN \cite{GCN} model, which is given by:
	\begin{equation} \label{equation_concat}
		{h_i} = \mathop {||}\limits_{s = 1}^S h_i^s
	\end{equation}
	where ${h_i} \in \mathbb{R}^{d_{final}}$ is the final embedding of node $v_i$, $d_{final} = H \times d_L \times |S|$, and $|S|$ is the number of strategies.
	\subsection{Classifier}
	In this work, we focus on the semi-supervised node classification task. We use a linear layer followed by a softmax to compute the predicted label probabilities, and employ the standard cross-entropy as the loss:
	\begin{equation} \label{equation_classifier}
		\begin{split}
			{\hat {y_i}} = \operatorname{softmax}&({h_i}\cdot W) \\ 
			L = -\frac{1}{{|{V_L}|}}\sum\limits_{{v_i} \in {V_L}}& {\sum\limits_{c = 1}^{|C|} {{Y_{ic}}\log ({{\hat y}_{ic}})} }
		\end{split} 
	\end{equation}
	where $W \in \mathbb{R}^{d_{final} \times |C|}$ is the learnable weight matrix and $Y_i ,\hat {y_i} \in \mathbb{R}^{|C|}$ is the one-hot embedding of label $y_i$ and the predicted label of $i$ respectively, $Y_{ic} = 1$ when $c = y_i$. The other elements in $Y_i$ are all set to zero.
	
	\section{Experiments}
	We now compare our RAW-GNN with the state-of-the-art models on the problems of node classification and visualization using seven real-world datasets varying from strong heterophily to strong homophily.
    
	\subsection{Datasets}
	To demonstrate that RAW-GNN can adaptively learn path propagation mechanism under both homophily and heterophily, we evaluate the performance of RAW-GNN and the existing state-of-the-art methods on seven real-world datasets, including three homophilic networks and four heterophilic networks. The features of these datasets are summarized in Table \ref{datasets}. $L.H.R.$ represents the label-defined edge homophily ratio of the network. $F.H.R$ represents the cosine feature edge homophily ratio of the network. 
	
	Cora, Citeseer and Pubmed are homophilic citation network benchmark datasets \cite{DatasetCite1,DatasetCite2}, where nodes represent papers, and edges represent citations between papers. Node features are the bag-of-words representation of papers, and node labels are academic topics. 
	
	Cornell, Texas and Wisconsin are webpage datasets collected from computer science departments of corresponding universities \cite{Geom-GCN}, where nodes represent web pages, edges are hyperlinks, node features are the bag-of-words representation of webpages, and node labels are pages categories (student, project, course, staff, and faculty). Actor is a heterophilic actor co-occurrence network \cite{DatasetFilm}, in which nodes correspond to an actor, and the edge between two nodes denotes co-occurrence on the same Wikipedia page. Node features correspond to some keywords in the Wikipedia pages. Node labels are categories in term of words of actor's Wikipedia.
	\begin{table}[ht]
		\centering
		\scriptsize
		\begin{tabular}{P{1.0cm}|*{6}{P{0.45cm}}P{0.55cm}}
			\specialrule{1pt}{0pt}{1.5pt}
			\textbf{Datasets} & \textbf{Texa.} & \textbf{Wisc.} & \textbf{Acto.} & \textbf{Corn.} & \textbf{Cite.} & \textbf{Pubm.} & \textbf{Cora} \\
			\specialrule{0.5pt}{0pt}{1pt}
			Nodes            & 183            & 251            & 7600          & 183            & 3327           & 19717          & 2708          \\
			Edges            & 309            & 499            & 33544         & 295            & 4732           & 44338          & 5429          \\
			Features         & 1703           & 1703           & 931           & 1703           & 3703           & 500            & 1433          \\
			Classes          & 5              & 5              & 5             & 5              & 6              & 3              & 7             \\
			$L.H.R$            & 0.06           & 0.18           & 0.22          & 0.30           & 0.74           & 0.80           & 0.81          \\
			$F.H.R$            & 0.35           & 0.34           & 0.16          & 0.31           & 0.19           & 0.27           & 0.17         \\
			\specialrule{1pt}{2pt}{1pt}               
		\end{tabular}
		\label{datasets}
		\caption{The Statistics of Datasets}
	\end{table}
	\subsection{Baselines}
	We compare our proposed approach RAW-GNN with the following baseline methods: 1) MLP (Multi-Layer Perceptron), which only uses node attributes; 2) Node2Vec \cite{Node2Vec}, which only uses graph structures. Since our work adopt a similar random walk sampling strategy of Node2Vec, we add it for comparison; 3) Traditional GNN models : GCN \cite{GCN} and GraphSAGE \cite{GraphSage}, which work under homophily assumption; 4) Frameworks designed for heterophily: H2GCN \cite{H2GCN}, CPGNN \cite{CPGNN}, GPR-GNN \cite{GPR-GNN}, BM-GCN \cite{BMGCN} and HOG-GCN \cite{HOGGCN}. In this paper, we choose the best results of each method for comparison, since some methods have more than one variants.
	
	\subsection{Parameter Setup}
	Following \cite{Geom-GCN} and \cite{HOGGCN}, we generate 10 random splits for all datasets. In each dataset, 48\% of the nodes are used as the training set, 32\% of the nodes are used as the validation set, and the rest as the test set.
	For a fair comparison, all methods use the same 10 random splits. All the parameters of the baseline methods were set as what were used by their authors. For the random walk sampling in RAW-GNN, we use DFS strategy with $p=10, q=0.1$ and BFS strategy with $p=0.1, q=10$. 
	We choose different path lengths from $\left\{ 3,4,5,6,7\right\}$  for different datasets.
	With every strategy, we sample 6 paths for each node in one epoch. For the RNN-based aggregator, we use GRU with 32 hidden units and attention head number is set to $2$. The learning rate is set to $0.05$. We adopt the Adam optimizer 
	and the default initialization in Pytorch.
	\begin{table*}[ht]
		\centering
		\scriptsize
		\begin{tabular}{P{2.3cm}|*{7}{P{1.4cm}}P{0.6cm}P{0.3cm}}
			\specialrule{1pt}{-1pt}{1pt}
			\textbf{Dataset} & \textbf{Texas}                  & \textbf{Wisconsin}              & \textbf{Actor}                    & \textbf{Cornell}                & \textbf{Citeseer}               & \textbf{Pubmed}                 & \textbf{Cora}     & \textbf{\tiny \begin{tabular}[c]{@{}l@{}}Avg\\Acc\end{tabular}} & \textbf{\tiny \begin{tabular}[c]{@{}l@{}}Avg\\Rank\end{tabular}} \\
			\specialrule{0.5pt}{-0.8pt}{1pt}
			MLP           & 83.24$\pm$5.77 & 86.47$\pm$3.33 & 36.49$\pm$1.16  & 84.32$\pm$6.14 & 69.10$\pm$2.69 & 86.37$\pm$0.64 & 72.98$\pm$2.31 & 74.14~                     & 6.00~                        \\
			Node2Vec(DFS) & 46.76$\pm$4.84 & 41.96$\pm$4.90 & 23.40$\pm$1.17  & 47.57$\pm$5.43 & 52.00$\pm$2.40 & 76.20$\pm$0.46 & 77.91$\pm$2.32 & 52.26~                     & 9.86~                        \\
			Node2Vec(BFS) & 44.05$\pm$6.29 & 40.39$\pm$5.28 & 23.33$\pm$0.96  & 45.95$\pm$8.29 & 45.80$\pm$2.62 & 65.76$\pm$0.44 & 72.03$\pm$1.86 & 48.19~                     & 11.00~                       \\
			GCN           & 55.68$\pm$9.61 & 53.73$\pm$7.65 & 30.64$\pm$1.49  & 55.14$\pm$7.57 & 74.81$\pm$1.87 & 87.25$\pm$0.56 & 86.60$\pm$1.44 & 63.41~                     & 8.29~                        \\
			GraphSAGE     & \underline{85.41$\pm$5.16} & 85.49$\pm$3.53 & 35.99$\pm$1.52  & 78.38$\pm$6.84 & 74.29$\pm$1.67 & 89.30$\pm$0.57 & 86.42$\pm$1.55 & 76.47~                     & 5.57~                        \\
			\specialrule{0.5pt}{1pt}{1pt}
			H2GCN         & 82.16$\pm$8.21 & 82.57$\pm$3.21 & 36.48$\pm$ 1.16 & 78.92$\pm$5.24 & \underline{75.95$\pm$2.18} & 88.78$\pm$0.53 & 87.69$\pm$1.37 & 76.08~                     & 5.00~                        \\
			CPGNN         & 74.32$\pm$7.38 & 81.76$\pm$6.74 & 35.51$\pm$1.85  & 63.51$\pm$5.83 & 75.52$\pm$1.84 & 89.08$\pm$0.67 & 87.18$\pm$1.13 & 72.41~                     & 6.29~                        \\
			GPR-GNN       & 84.59$\pm$4.37 & 83.92$\pm$3.14 & 36.47$\pm$1.38  & 82.97$\pm$5.68 & 75.12$\pm$1.98 & 87.38$\pm$0.63 & 86.70$\pm$1.03 & 76.74~                     & 5.43~                        \\
			BM-GCN        & 85.13$\pm$4.64 & 82.82$\pm$8.89 & 36.32$\pm$1.35  & 79.14$\pm$8.44 & 75.94$\pm$2.36 & \textbf{90.25$\pm$0.71} & \underline{87.71$\pm$1.11} & 76.76~                     & 3.86~                        \\
			HOG-GCN       & 85.17$\pm$4.40 & \underline{86.67$\pm$3.36} & \underline{36.82$\pm$0.84}  & \underline{84.32$\pm$4.32} & \textbf{76.15$\pm$1.88} & 88.79$\pm$0.40 & 87.04$\pm$1.10 & \underline{77.85~}                     & \underline{2.86~}                        \\
			\specialrule{0.5pt}{0.8pt}{1pt}
			RAW-GNN        & \textbf{85.95$\pm$4.15(4)} & \textbf{88.24$\pm$3.72(4)} & \textbf{37.45$\pm$1.21(5)}  & \textbf{84.86$\pm$5.43(4)} & 75.38$\pm$1.68(5) & \underline{89.34$\pm$0.66(4)} & \textbf{87.89$\pm$1.52(7)} & \textbf{78.44~}                     & \textbf{1.71~}                       
			\\
			\specialrule{1pt}{0.5pt}{0pt}
		\end{tabular}
		\label{classification_accs}
		\caption{Classification Results with mean value and standard deviation. The best result is bold and the second best is underlined. For our RAW-GNN, the number after deviation denotes the random walk path length chosen for the corresponding dataset.}
	\end{table*}

	\begin{figure*}[ht]
		\centering
		
		\subfigure[GCN] {  
			\includegraphics[width=0.30\columnwidth]{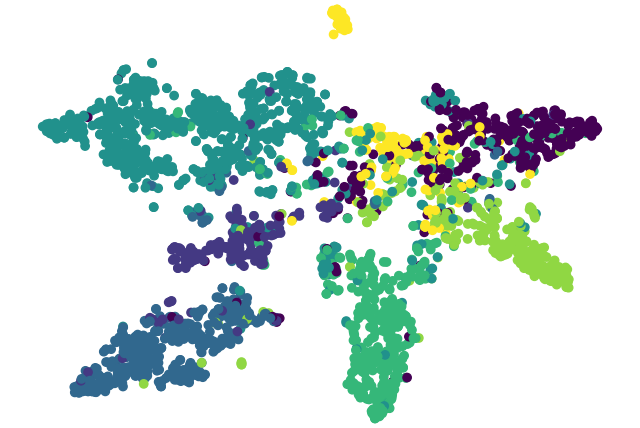}
		}
		\subfigure[CPGNN] {
			\includegraphics[width=0.30\columnwidth]{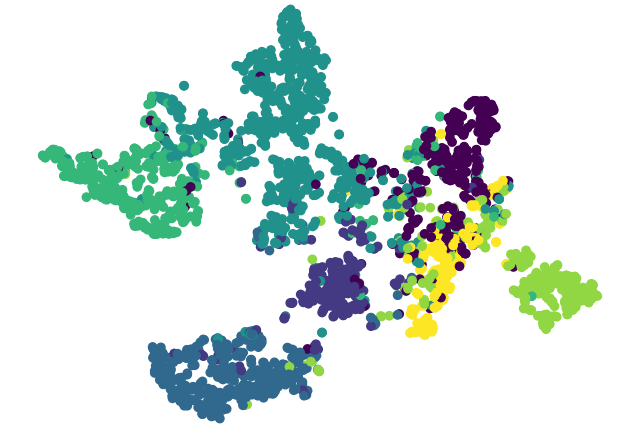}
		}
		\label{attcc}
		\subfigure[H2GCN] {
			\includegraphics[width=0.30\columnwidth]{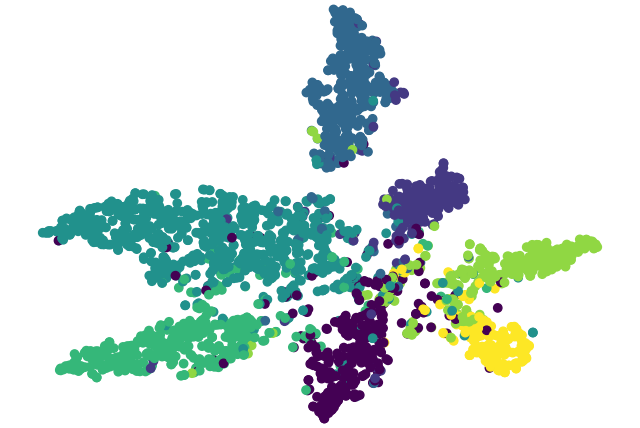}
		}
		\label{attbb}
		\subfigure[GPRGNN] {
			\includegraphics[width=0.30\columnwidth]{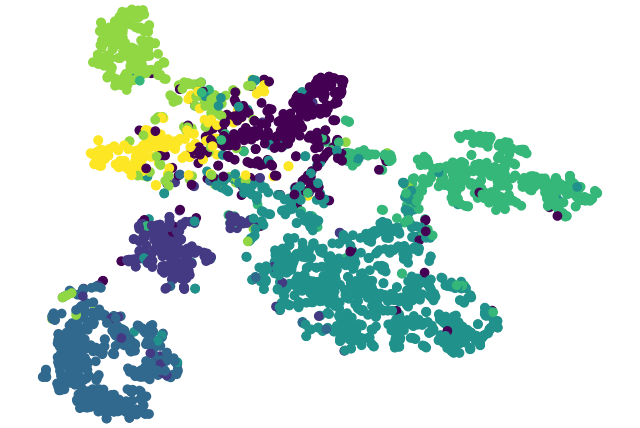}
			
		}
		\subfigure[RAWGNN] {
			\includegraphics[width=0.30\columnwidth]{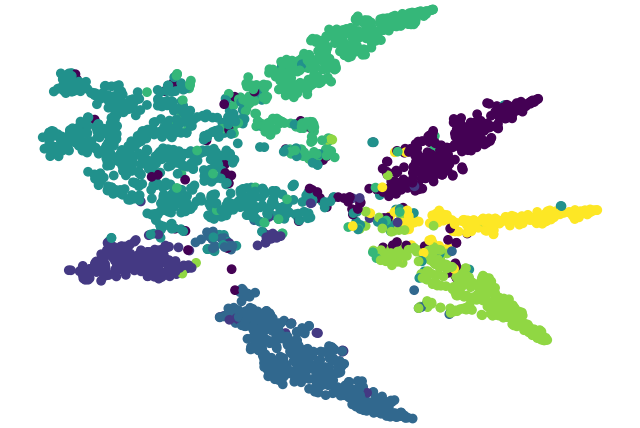}
			
		}
		\caption{Visualization results on Cora dataset}
		\label{visualization}
	\end{figure*}
	
	\subsection{Node Classification}
	We conduct experiments on seven real-world datasets to compare the performance of different models for node classification (Table \ref{classification_accs}). We use the mean accuracy and standard deviation as the evaluation metric. As shown, our RAW-GNN performs best on 5 out of 7 networks. Below are the detailed observations.
	\begin{itemize}
	    \item  Our RAW-GNN approach performs the best on all of the four heterophilic networks, i.e., Texas, Wisconsin, Actor, and Cornell, which empirically proves the effectiveness of RAW-GNN. To be specific, RAW-GNN outperforms heterophily-agnostic models, i.e., Node2Vec, GCN and GraphSAGE, by 34.20\%, 25.33\% and 2.81\% respectively. This is largely because the rival methods cannot generalize to heterophilic scenarios. GraphSAGE performs relatively well and we assume the reason is the use of neighbor sampling like our RAW-GNN, which demonstrates the effectiveness of sampling in heterophily graphs. 
	    Compared with the methods designed for heterophily, such as H2GCN, CPGNN and GPR-GNN, BM-GCN, and HOG-GCN,  RAW-GNN also achieves an improvement between 0.88\% and 10.35\% by mean accuracy. These results demonstrate the effectiveness of our new method on heterophily networks. 
	    
		\item On homophilic networks, i.e., Cora, Citeseer, and Pubmed, RAW-GNN also has competitive performance compared with the baselines. Specifically, RAW-GNN performs the best on Cora and the second best on Pubmed. We assume that since every dataset is a mixture of homophily and heterophily, the heterophily information from the DFS channel also helps model performance on these homophily datasets. Note that RAW-GNN outperforms GCN and GraphSAGE on all these datasets by 1.32\% and 0.87\% by mean accuracy. These results demonstrate that RAW-GNN still maintains comparable performance on homophilic datasets. Furthermore, for datasets with lower cosine feature homophily ratios ($F.H.R < 0.2$), the best path-length is longer, which indicates that heterophily networks need large receptive field to extract the hidden neighborhood distribution in the network. These results show the effectiveness and robustness of the proposed framework empirically.
	\end{itemize}
	
	\subsection{Visualization}
	In addition to the quantitative node classification, we also visualize node embeddings on Cora dataset to assess the embedding results qualitatively. We project the node embeddings into a $2$-dimensional space using t-SNE \cite{t-SNE}. Here we illustrate the visualization results of GCN, H2GCN, CPGNN, GPR-GNN and our RAW-GNN in Fig. \ref{visualization}, where points with different color indicate different classes.
	As shown, the visualization results of GCN and CPGNN are less satisfactory in this case, since points with the same class are dispersed and some points with different classes are mixed. As shown, the visualization of GPR-GNN and H2GCN are better but still blurred along the border of different classes. The result of our RAW-GNN is the best, where the border between different classes is the most discernible. This result is consistent with classification result.
	
	\section{Conclusion}
	In this paper, we propose a random walk aggregation based graph neural network that can process homophily and heterophily graphs at the same time. The proposed framework extends the neighborhoods in traditional GCNs to k-hop path-based neighborhoods generated by two random walk sampling strategies (i.e., breadth-first search and depth-first search). Then, the proposed framework uses the sequential RNN-based aggregator to encode the ordered attribute information of neighbor nodes. Then, the path embeddings for each strategy are gathered to the target node with an attention mechanism to form strategy-specific embedding. At last, node embeddings from different strategy channels are concatenated to prevent information of different characteristics from mixing and enable the model to automatically trade-off between homophily and heterophily according to different network characteristics. Experiments on seven real-world datasets demonstrate that the proposed approach outperforms existing methods under heterophily, and also performs competitively under homophily. 
    \section*{Acknowledgments}
	
	This research was partly supported by the Natural Science Foundation of China under grants 61876128, the Tianjin Municipal Science and Technology Project (Grant No. 19ZXZNGX00030), and Meituan Project.
	\bibliographystyle{named}
	\bibliography{sample-base}
	
\end{document}